# MBSE analysis for energy sustainability improvement in manufacturing industry

Romain Delabeye
Quartz Laboratory, EA7393
ISAE-Supméca,
Saint-Ouen, France
0000-0001-7873-2904

Olivia Penas
Quartz Laboratory, EA7393
ISAE-Supméca,
Saint-Ouen, France
0000-0002-6363-5754

Martin Ghienne
Quartz Laboratory, EA7393
ISAE-Supméca,
Saint-Ouen, France
0000-0002-6315-538X

Arkadiusz Kosecki
Quartz Laboratory, EA7393
ISAE-Supméca
Saint-Ouen, France
0000-0002-6659-7044

Jean-Luc Dion
Quartz Laboratory, EA7393
ISAE-Supméca
Saint-Ouen, France
0000-0002-0124-9030

*Abstract* — With the ever increasing complexity of Industry 4.0 systems, plant energy management systems developed to improve energy sustainability become equally complex. Based on a Model-Based Systems Engineering analysis, this paper aims to provide a general approach to perform holistic development of an autonomous energy management system for manufacturing industries. This Energy Management System (EMS) will be capable of continuously improving its ability to assess, predict, and act, in order to improve by monitoring and controlling the energy sustainability of manufacturing systems. The approach was implemented with the System Modeling Language (SysML).

## I. Introduction

In line with the ambitious objectives of the Paris Agreement and Sustainable Development Goals [1], the current global awareness of environmental risks makes the issue of sustainability essential for anyone. Indeed, sustainable development meets the needs of the present without compromising the ability of future generations to meet their own needs [2]. In this respect, energy sustainability – as it refers to the provision and use of energy services in a sustainable manner, all along the related lifecycle – offers a fundamental driver to contribute to the reduction of the related environmental, economic and social impacts [3]. It relies on three key pillars: sustainable energy supply, sustainable energy consumption and sustainable waste disposal [4].

The resultant change in manufacturing companies requires them to adapt both their organization and manufacturing strategies in order to cope with these new challenges [5]. Indeed, although they are not yet driven by sustainability factors, the production function is a key driver in a company's strategy to tackle environmental concerns. Whether it concerns the sustainable supply of energy (energy sources and carriers, including their transport and storage), the sustainable consumption of energy or the sustainable disposal of waste, companies have to pay more attention to their manufacturing footprint, especially as energy-related costs are becoming increasingly high [6]. In this context, improving understanding of the energy sustainability challenges and solutions is therefore becoming of critical importance for manufacturing industry decision makers, which need to gain a comprehensive knowledge of how to evaluate, visualize and manage a production system in order to reduce its environmental impact while contributing to the competitiveness of their company.

As the result, the EU-funded EnerMan (Energy-efficient Manufacturing system management) project (2021-2024) intends to improve energy efficiency by monitoring energy-related flows and data. This project will provide a platform, running over the entire life cycle of a manufacturing plant, which is capable of adapting to predefined energy sustainability indicators, predicting energy consumption during changes impacting current and future production or fluctuations in external costs related to the market for the energy supplied. Besides, it will then provide autonomous updates of plant process control, production lines and equipment, while training plant operators and decision makers in energy sustainability through best practices. To this end, this platform (hereafter named: Energy Management System (EMS)) will rely on a systemic and holistic approach, collecting and processing real data from multiple manufacturing sources, and then generating simulation data to extract energy prediction results. Finally this EMS will be able to assess the energy sustainability of a plant, at its different scales, and to autonomously and flexibly adapt the control management of its different manufacturing processes in a cognitive way, in order to minimize its systemic energy consumption and reduce its environmental footprint.

The objective of this paper is to provide a Model-Based System Engineering (MBSE) analysis to support the development of a systemic and holistic approach aiming at improving energy sustainability in the manufacturing industry. Section 2 will present the state-of-the-art related to energy sustainability representations and related frameworks. Section 3 will detail the MBSE analysis developed. Section 4 proposes a discussion before the conclusions and perspectives (Section 5).

## II. State-Of-The-Art

### A. Energy sustainability representation

Energy sustainability modeling needs to take into account many interrelated processes. A classic approach to address this problem is to represent the set of levels of the production system, from the machines' components to the entire factory [7]. An energy-efficiency analysis plan was developed in [4], with several point of views: device and process levels; line, cell and multi-machine systems; facility and multi-factory systems; enterprise and global supply chain. Indeed, the component-to-machine levels may focus for example on control and machining process enhancement, while higher levels, e.g. the production line or the plant, often tackle process optimization and scheduling. Energy sustainability modeling with multi-scale representations is also common in strategic energy planning for large-scale energy systems from smart cities to national policies[8], [9]. For other energy-intensive systems such as data centers [10] and smart buildings [11], physics-based spatial representations are sometimes preferred, e.g. using Computational Fluid Dynamics (CFD) for temperature management and server cooling.

The representations above, from which application-specific models and meta-models can be developed, are useful for extracting and visualizing data, or solving complex technical problems. These can inter alia support energy sustainability parameters estimation and metrics evaluation, e.g. the influential factors mentioned in the ISO 14955 [12] or ISO 50006 [13] norms, as well as many other parameters [14]–[19]. When designing a complete EMS, several representations among the ones cited can co-exist to solve dedicated problems or satisfy different stakeholders. For example, focusing on the process optimization, unified representations have been proposed, namely hybrid models [20], consisting in a discrete state diagram with multiple dynamical systems and modes. However, none of these previous representations allows for a consistent integration of all the underlying patterns between remotely correlated parameters or metrics, to address the numerous issues related to energy sustainability, hence requiring a general framework.

Authors of [4] proposed a framework in which each energy sustainability enhancement is classified according to the lifecycle phase of the energy it treats: energy supply, consumption or disposal. In parallel, Rosen proposes in [2] a global analysis according to its impacts type: social, economic or environmental; and a classification into the following categories: energy efficiency, e.g. energy loss and waste reduction, or energy effectiveness, e.g. energy needs reduction and lean production. Still, these frameworks still lack of methods to identify the interplays between the different enhancement proposals and the system behavior, parameters and metrics. Moreover, the modeling of external constraints may present an additional difficulty.

Finally, some commercial EMS are currently being developed [21]–[24], but no associated research studies have, to our knowledge, proven their capabilities regarding autonomous control, learning continuous improvement, adaptability to legacy manufacturing systems, in accordance with the ambitions of the EnerMan project.

*B. Complex system frameworks*

Even though the aforementioned representations can serve as bases for problem solving to improve energy sustainability, the development of an intelligent and autonomous EMS, relying on the needs and contributions of the numerous stakeholders, requires to be treated as a complex system, to ensure its feasibility and consistency.

Regarding existing complex system frameworks, Interpretive Structural Modeling (ISM) methods [25] can decompose problems and structure their resolution. This framework allows for knowledge management [26], [27] and possesses a certain versatility and simplicity of use. This may clarify the problems to overcome, but would not be sufficient to accurately represent the system's internal and external interactions. Similarly, numerous methods divide the original problem into tractable issues to address, e.g. by means of functional flow decomposition [28], [29], enterprise architecture frameworks for a certain class of organizational issues [30].

In parallel, System Engineering community [31]–[33] provides numerous methods, tools and guidelines to support the specification, analysis et development of any complex system, while ensuring in the MBSE context the consistency and traceability of any modeling artefacts, valuable for managing numerous heterogeneous interactions. Indeed MBSE methods can meticulously identify the stakeholders' needs and contributions, integrate system constraints, in order to define various system architecture, while facilitating the verification and validation process [34], [35].

In conclusion on the modeling approaches of complex systems adapted to address energy sustainability, MBSE fulfills many advantages, which however have not been exploited to date to address the issues of improving the energy sustainability of manufacturing systems, while enabling the integration of specific representations previously identified to address some local interactions between elements of MBSE models.

III. MBSE ANALYSIS

Improving energy sustainability is an important part of sustainable manufacturing. This is a wide topic though, with numerous stakeholders as well as a myriad of interconnected research topics and solutions of different nature. As a result, this issue possesses the characteristics of a complex system, which makes MBSE a suitable framework for decomposing and analyzing sub-systems interactions, in order to support the development of a systemic and holistic approach fulfilling the numerous related requirements. The MBSE model proposed herein has been developed in the Systems Modeling Language (SysML), using the Cameo software v.19.0.

The workflow of the developed MBSE analysis is based on four main top-down steps: (i) the derivation of the energy sustainability requirements limited to the scope and constraints of the EnerMan project; (ii) an analysis of the macro-functions (corresponding to the different energy sustainability improvement paths) that the energy management system (EMS) to be developed must have in order to meet these requirements; (iii) the scientific research areas that can contribute to these macro-functions; (iv) the solutions/technologies for the implementation of these approaches. Finally, the traceability of the dependencies between these four views, conducted all along this workflow, allows to identify during the project, for a given modeling element (e.g. treated by a partner within the project), all the ins and outs and in particular the interfaces to be foreseen and defined with other interacting parties, but also to facilitate the crucial steps of verification and validation.

## A. Requirements specification

Taking "Improve energy sustainability in a manufacturing company's context" as the main requirement, a first decomposition highlights two key ones "Reduce sustainability impacts" and "Improve overall energy efficiency" (Figure 1).

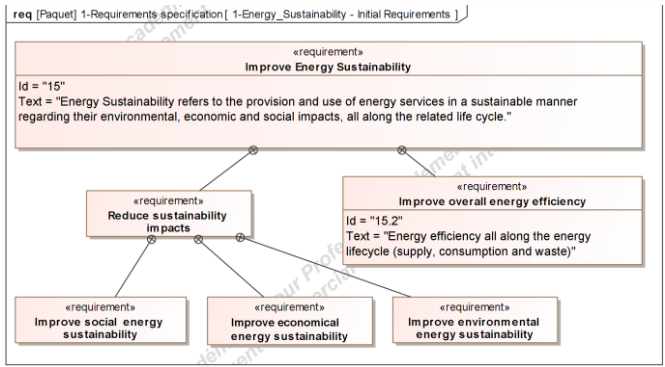

*Figure 1: Initial Requirements*

At this step of initial requirements, some additional requirements can be derived, such as legislation, standards and label compliance, but also Corporate Social Responsibility policies [36] and sustainability awareness (for staff and to follow worldwide trends) for social impacts as well as economic strategies, e.g. green investment decision-making [37] for economic impacts.

Focusing on the perimeter of the EnerMan project, we analyzed its context, identifying all the stakeholders, their expectations and their contributions (Figure 2).

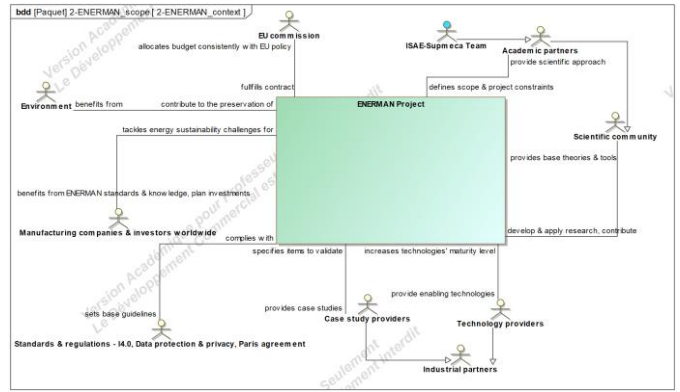

*Figure 2: EnerMan project context*

Then, we have detailed the requirements related to the EMS to be developed and expected from the EnerMan research work. To this respect, we have decomposed the main objectives of the project in relation with the previous energy sustainability-related requirements (Figure 3), then we have derived these main goals to elicit all the requirements that must be met by the project deliverable (i.e. EMS), as shown in Figure 3.

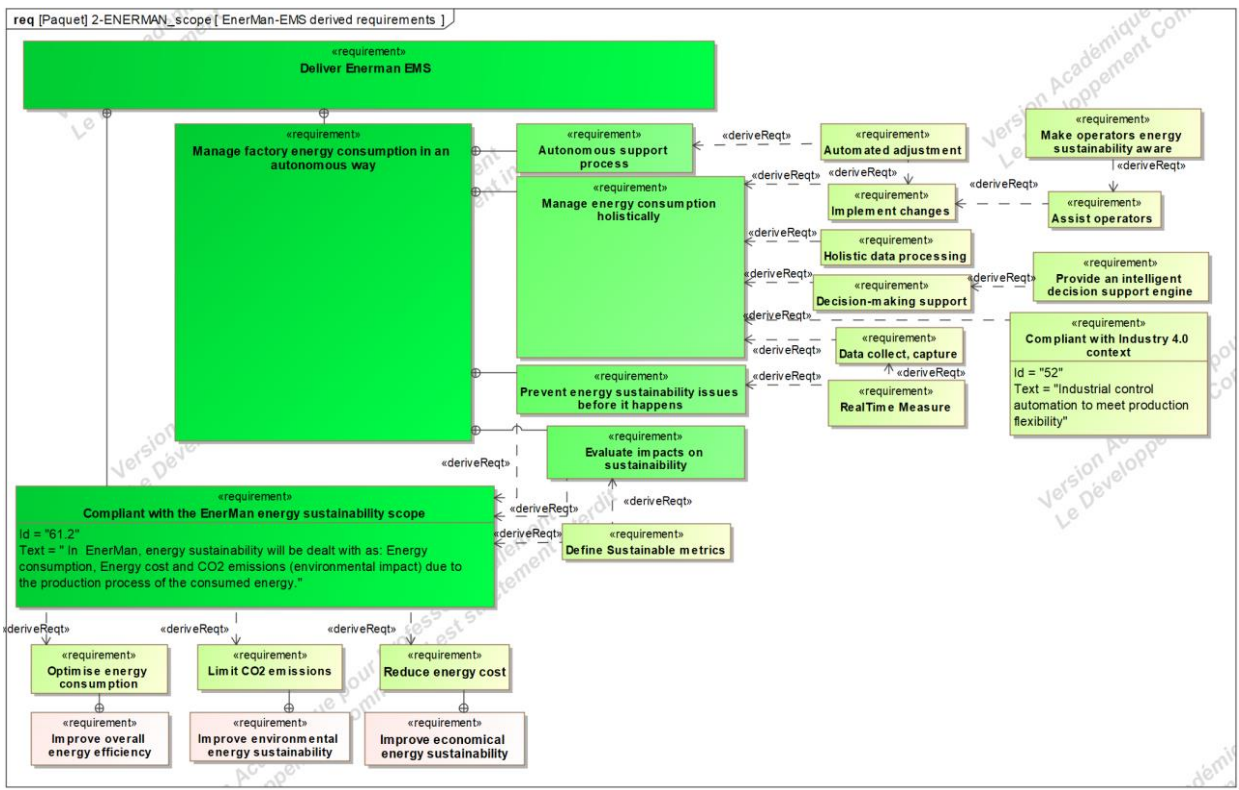

*Figure 3: Extract of the EMS derived requirements*

In the EnerMan scope, the main requirement deals with the delivery of an EMS, which consists in being compliant with the EnerMan energy sustainability scope while managing factory energy consumption in an autonomous way. Regarding the first aspect, the emphasis is on tackling energy sustainability, defined as the coincidence of energy consumption and cost optimization, as well as $CO_2$ emission reduction, in line with the initial requirements for improving energy sustainability. Furthermore, it should be noted that most production processes are interconnected, not only from the process point of view but also from the impact of the machines on their common environment. The energy consumption of a plant must therefore be treated holistically rather than as the uncorrelated sum of its components. This leads to strong constraints in terms of data collection, data processing and decision making, but can be extended to cyber-security measures. Another requirement addresses the prognosis needs, to prevent energy sustainability issues before they happen, then bringing the time for energy analysis and control to a near real-time level, and requiring the evaluation of the various energy sustainability impacts.

These first level requirements are the result of a preliminary analysis aiming at reducing the global complexity of the problem by structuring the different levels and dependencies, in order to facilitate later the verification and validation processes. Therefore, they already guide the definition of the system architectures that will allow the development of the final product.

### B. EMS Functional Architecture

The functional architecture of the EMS has been built from the previous requirements and after having analyzed the life cycle of the legacy production system (LPS). Indeed, one of the important and ambitious constraints of the addressed project is to develop an EMS that can be adapted to the majority of LPS, as an add-on system, in order to allow easy operation and deployment into many European manufacturing industries. Thus, the analysis of the context and use-cases diagrams of the EMS according to the phases of the LPS lifecycle allowed to represent the expectations of the stakeholders and to bring out the macro-functions, with respect to the LPS operating modes during each of these phases. In addition, these diagrams highlighted the external interfaces of the EMS that support the exchange of physical or information flows with other stakeholders, and in particular the LPS (Figure 4).

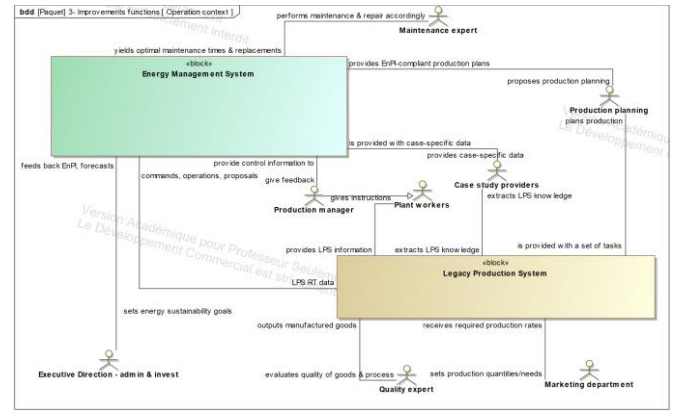

*Figure 4: EMS Context for the LPS operating phase*

Then the EMS functional architecture has been defined in SysML activity diagrams, with a top-down decomposition from the highest level defining the macro-functions, including their physical, information, command and energy interactions (Figure 5). External interfaces have mainly been named with the acronym related to its source (external actor or internal function), to facilitate the understanding (LPS X coming from LPS, EMS X coming from an internal function, etc.

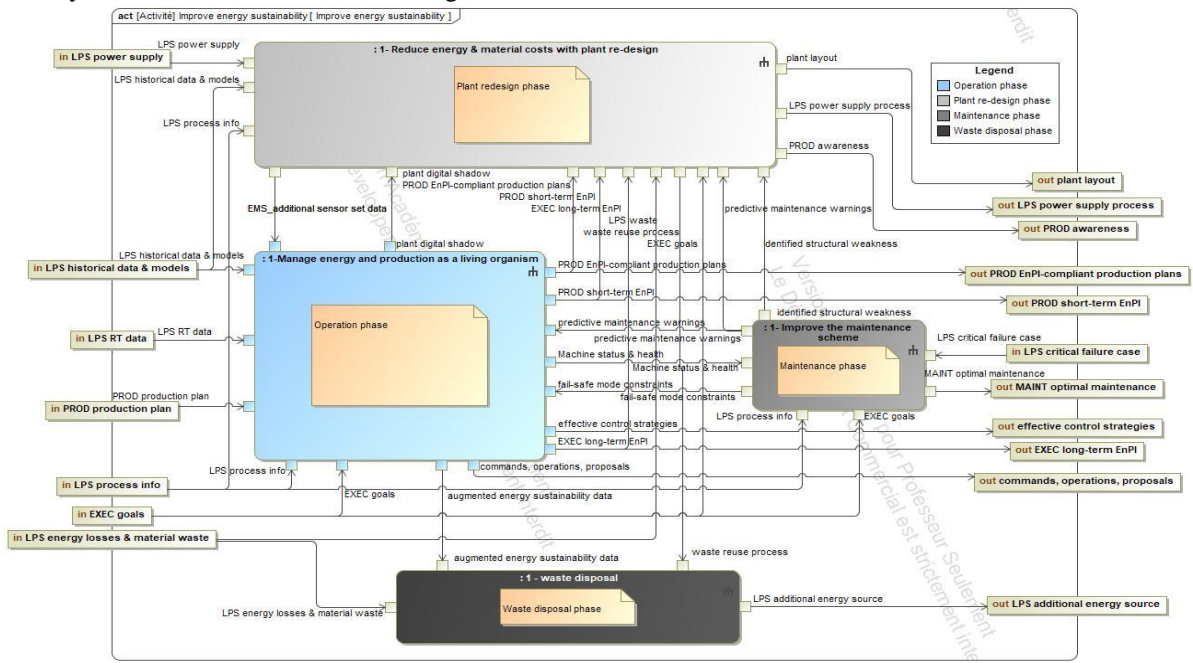

*Figure 5: EMS functional architecture with respect to the Legacy Production System lifecycle phases*

These system functions have been modeled to different extents, proposing thus sub-functions to address with a view to enhance manufacturing plants energy sustainability for each LPS lifecycle phase, namely the operation, maintenance, waste disposal and plant re-design phases. As a base hypothesis, considered LPS has already been set up and is running. In the operation phase, the EMS has to perform three main functions respectively related to monitoring, analysis and control tasks (Figure 6).

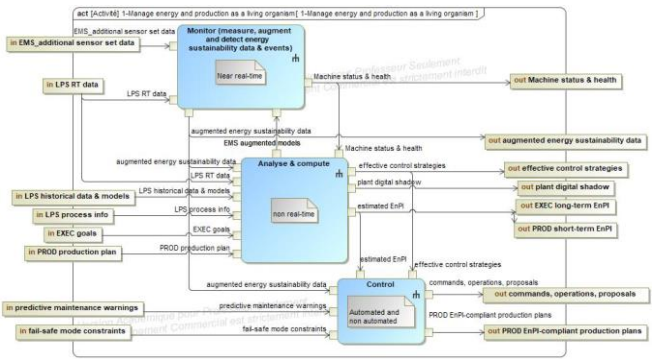

*Figure 6: Focus on EMS functions in operation phase*

The first sub-function (focused on monitoring) is then subdivided into energy monitoring and condition-based monitoring. The latter is the angular stone of predictive maintenance, where machine health has to be constantly monitored, while the former estimates LPS energy-sustainability status in real-time. Both sub-functions are correlated, as an energy drift or event may be an evidence of component failure, and vice versa. The second sub-function (analysis) is crucial in that it implies simulation and prediction capabilities, allowing for complete energy sustainability performance assessment. This assessment may include, but is not limited to, regularly updated life cycle assessment (LCA) [38] or cost (LCC) [39], and short and long term energy sustainability performance metrics.

The maintenance phase mainly addresses component replacement and optimal maintenance planning, preferably according to an energy-sustainable cost function. Curative, palliative and predictive maintenance activities have been distinguished, as these could be performed by different scientific approaches and physical solutions, as described in the next sections. Depending on the maintenance diagnoses or prognoses, LPS structural weaknesses may be observed and call for their re-design. Typically, this re-design phase is responsible for proposing new processes that enhance the waste disposal, the power supply (to be adapted or supplemented by local energy production systems), as well as the energy efficiency of LPS equipment. This can be done either by enhancing plant layout, designing more energy-efficient production lines, or providing trainings and instructions to the personnel, thus increasing personnel energy-sustainability awareness.

Eventually, unrepairable or obsolete equipment, as well as energy or material waste capable of being harvested, will be recycled and injected back into the operation phase.

Finally, this multi-scale functional representation allows to identify all the interactions between the different functions and then to highlight the interfaces and dependencies as respective function ins and outs. As a result, when focusing on a given EMS function, whatever its level, it is possible to identify which other sub-functions will provide inputs including their (physical) forms, and similarly those that will receive the outputs generated.

Figure 7 allows, as an illustrative example, to analyze the direct inputs and outputs of the sensor fusion sub-function "Local data fusion", derived from the "Energy monitoring" function. Thus, this sub-function aggregates many data that must be easily available, with an appropriate specific sampling rate: the EMS augmented models come from the models generated by the internal "Analysis and compute" function of the N+2 level (Figure 6), the sensor data (initial from the LPS or those added if necessary by the EMS redesign function) and the augmented (extracted from the characteristics) and energy data. These interfaces with other functions (of the same level or not) lead to new (induced) requirements that will allow, thanks to quantitative refined requirements, to establish holistic verification criteria of all functions.

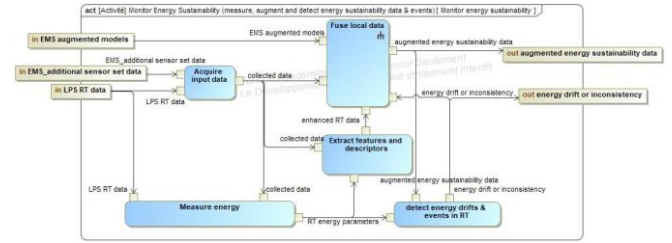

*Figure 7: Monitor Energy Sustainability low-level function*

### C. Research-focused architecture

The objective of this step is to identify the relevant scientific approaches that can be performed to deal with the functions previously described. They have been synthesized and organized into a Block Definition Diagram (BDD), with their hierarchical dependencies (composition and shared aggregation, depending on their involvement and specialization degree) in the Figure 8.

Based on the EMS requirements for intelligent (self-adapting) autonomous management of energy sustainability, including real-time data from the LPS, the digital twin approach emerges as the primary scientific approach to support the overall EMS design. Its main components are operations research, quality control, cyber security, modeling approaches, and physical integration. Artificial intelligence and optimization techniques are ubiquitous in the scientific approaches considered, and often work in combination, therefore they are also part of the digital twin approach. This architecture has been decomposed at different levels of detail, in accordance with the functions identified in the previous step. In particular, the modeling approaches can be very different depending on the end use. For example, modeling for control (including sensor fusion models for state estimation) can be approximate models, linear or linearized approximations (when possible), where the model is regularly recalculated and the need for high fidelity models thus reduced (especially when the characteristic time of the system dynamics is much slower than the control frequency); whereas models for system identification often aimed at characterizing and analyzing the system dynamics, which requires high fidelity. This class of models will typically use regression techniques, together with model order reduction and potentially interpretable descriptive models.

Compressed sensing is another modeling class, an enabler for technological integration which can complement the other models by specifying where to place the sensors and which information to retrieve in order to generate or calibrate accurately-enough data-driven models. Finally, predictive maintenance approaches such as fault detection and isolation also play a critical role in enhancing energy sustainability; these methods are integrated into the modeling and processing approaches.

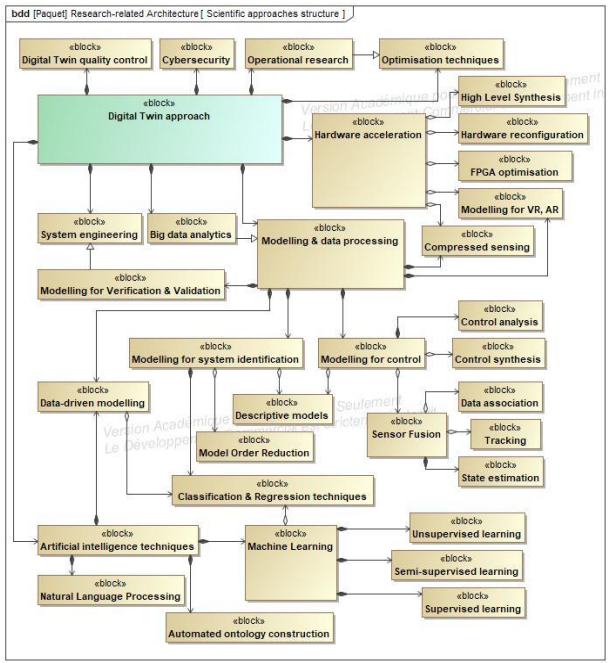

*Figure 8: Scientific approaches structure to support the EMS development*

The allocation links between the functions (activities) and these related scientific approaches have been declared in a SysML allocation matrix. This matrix allows hence for defining the nature of the technical exchanges to have with other fields of research and project stakeholders, with a view to complete the targeted functions. From a scientific perspective, the research fields are increasingly sharing and co-developing methods to solve modern complex problems such as sustainability. For instance, while machining energy efficiency has been extensively addressed by diverse scientific communities, reducing machining energy consumption at plant level will typically combine production management methods, e.g. operational research approaches, with automation methods such as advanced control approaches. However, for reasons of limited length of the paper, we have chosen to focus here on the analysis of the functions belonging to our research scope. Therefore, the corresponding allocation matrix regarding our research scope is presented in Figure 9.

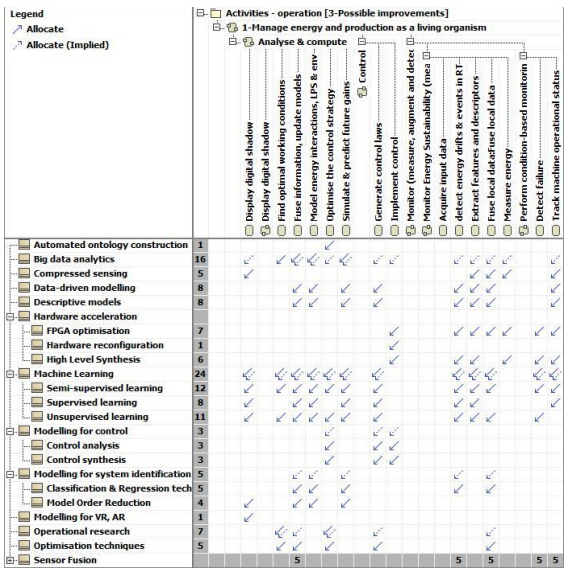

*Figure 9: Extract of the EMS allocation matrix*

This step is crucial to define alternative scientific approaches to develop the different functions, and to identify possible combinations of approaches within the same function, and thus initiate scientific collaborations to benchmark the performance of different scientific approaches around the same function or to jointly develop hybrid approaches.

### D. Technological architecture

In this last step, we present alternative technological solutions for the physical implementation (hardware and software) of the scientific approaches previously identified. We have structured this architecture in 3 main technological subsystems (Figure 10): the Intelligent Cyber Physical System (CPS) End Nodes, where LPS data are gathered and pre-processed for automation purposes to be further analyzed and augmented within the Analysis Computing Center including prediction and simulation capabilities in order to feed the Intelligent Decision Support System in charge of determining suitable energy sustainability control strategies to be returned to the Intelligent CPS End Nodes.

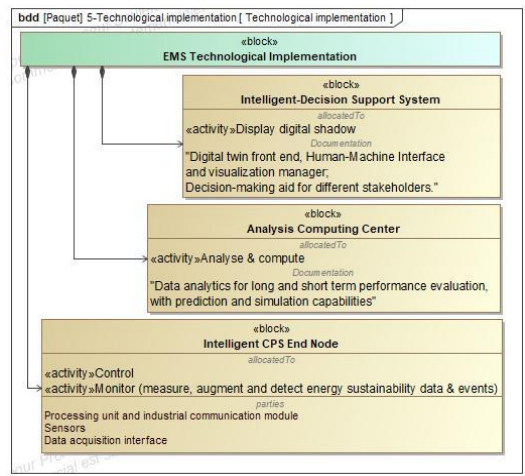

*Figure 10: EMS physical implementation*

Focusing on the Intelligent CPS End Nodes (Figure 11), it is composed of sensors, data acquisition and low power processing and communication units, typically reconfigurable FPGA boards.

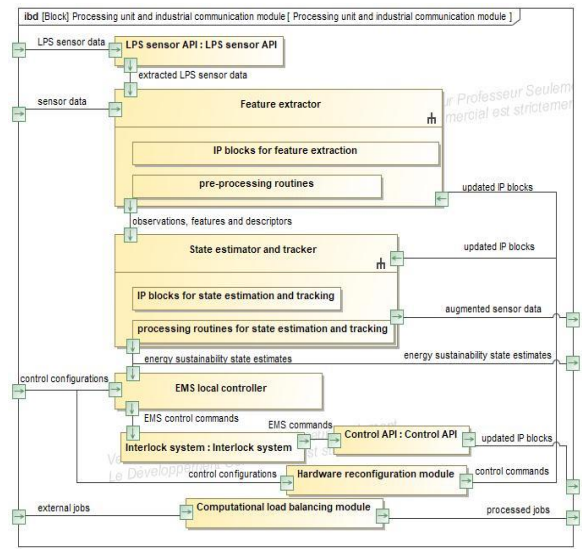

*Figure 11: Intelligent CPS End Nodes software architecture*

An interlock system must also be present for safety reasons. Software-wise, the "Processing unit and industrial communication module" sub-component will integrate a low-level feature extractor, yielding for instance statistical and spectral descriptors, a state estimator, e.g. tracking energy sustainability states of interest even when these are not observable, complementing a controller.

As above, a traceability matrix finally provides the correspondence between the defined scientific approaches and these technological solutions, using allocation links.

## IV. DISCUSSIONS

The general MBSE approach previously detailed is a preliminary research work aiming at supporting EMS design to improve energy sustainability of manufacturing industry, by providing an improvement process capable of adapting to real-time events of the machines and production lines, to achieve better performances. It is based on a consistent holistic model describing and connecting the stakeholders' needs, the derived requirements, the corresponding system functions, but also in a more original way the relevant scientific approaches and associated technologies to support their implementation on existing industrial systems. Thus, this model can serve as a basis for many scientific and industrial communities that would like to address the same problem (energy sustainability) with different scientific approaches and technical solutions. Besides a scientific state of the art and a technology watch on the relevant technical solutions that can answer the issue of energy sustainability, our contribution also addresses the relevance of these techniques with respect to the specific constraints of the industrial scenario (objective, priority, feasibility) but also to the constraints of collaboration imposed by a collective work between various partners. Indeed, as the project is composed of partners with little to no previous joint research or industrial activities, the sustainability enhancement solutions proposed by the partners will result from initiatives and yet will have to be developed and integrated together at the end with the same EMS. The holistic view of the proposed model allows each partner to position their research and corresponding implementations, while being aware of the interface or compatibility constraints imposed by the development of other contributors' modules. Finally, our work contributes to support the collaborative design of an EMS for the improvement of energy sustainability, while offering the flexibility and agility necessary for the diversity of the industrial cases studied and the partners involved.

In addition, this model is useful for identifying the ins and outs of our research area with respect to the scientific issues raised. Thus, taking into account our role within the project, the expertise of our research team and the technical and temporal constraints of our development environment, we are able to define the perimeter of our forthcoming scientific contribution from the functional level, the possible scientific approaches to address the identified functions up to the associated technological solutions for the physical implementation of the EMS (Figure 12). The traceability links of these different elements with the initial requirements of the EMS will allow us to clarify the validation procedures, while those addressing the derived requirements limited to the EnerMan perimeter will support the verification stage of our future results with respect to the project objectives.

As a consequence, our upcoming scientific contributions will target the LPS operation phase, and more specifically the real-time monitoring and control functions. We will design solutions so that the EMS is able to perform the following key sub-functions: measure energy, extract features and descriptors (from available energy sustainability data), fuse local data, and generate control laws. The identified scientific approaches that enable these activities to be carried out will include compressed sensing, modeling for control, modeling for system identification and sensor fusion (Figure 12), with underlying machine learning techniques. The technological solutions developed from the approaches mentioned will be physically implemented into the Intelligent CPS End Node, thus contributing to the main edge software components, i.e. the feature extractor, the state estimator and tracker, as well as the EMS local controller. Hardware-wise, those software components will either be made to run on FPGA boards or embedded processing units.

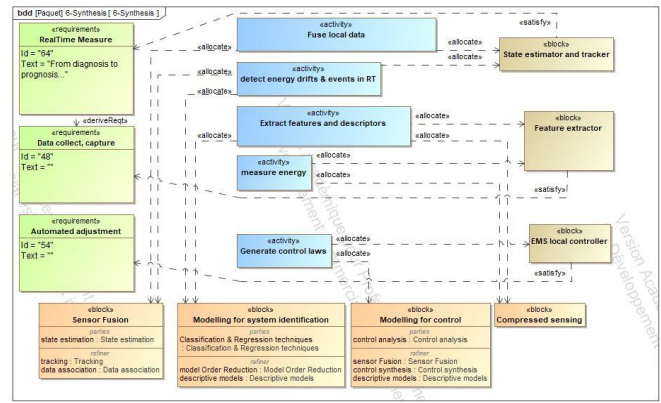

*Figure 12: MBSE model synthesis*

The models proposed in this paper have been structured so as to ease the positioning and integration of a stakeholder's contribution within the overall scheme. Shared and cross validation requirements ensure the contributions' consistency and effectiveness, with traceability considerations at each step of the design process.

Finally, the EMS presented herein does not presume the final EnerMan system structure, as this shall be the result of collegial decisions at consortium level. Such an approach is complementary though, and can lead to original technical proposals to address the energy sustainability subject in the manufacturing industry context. Not only this MBSE approach has the ability to model the inherent interplays within the EMS and with respect to the production system, these allow for identifying common interests and research activities, between the stakeholders.

## V. CONCLUSIONS & PERSPECTIVES

The MBSE analysis developed in SysML allows to capture, in an exhaustive and coherent way, the complex aspects of energy sustainability for the manufacturing industry, to facilitate the design of an intelligent, autonomous and self-adapting energy management system acting on a production system. The originality of this work also lies in the exploration of scientific approaches and technological solutions to address the different functions of this EMS, as well as the definition of their dependencies. Moreover, this research work allowed us to highlight our scientific positioning in relation to this issue, taking into account our research and technical constraints. Future work will focus on the implementation of this EMS within the scope of our research area. As such, this may include the refinement of requirements for each subsystem, keeping the holistic vision. Finally, the interactions between the EMS and its environment will also be further detailed, both to better capture the interactions between the factory's machines, and to dynamically mitigate the negative impact of external hazards, on the factory's energy efficiency.

ACKNOWLEDGMENT

EnerMan has received funding from the European Union's Horizon 2020 research and innovation programme under grant agreement No 958478.